\documentclass{IEEEtran}

\usepackage{cite} 
\usepackage{epsfig}
\usepackage{subfigure}
\usepackage{url}
\usepackage{amssymb}
\usepackage{amsmath}

\newtheorem{theorem}{Theorem}[section]
\newtheorem{lemma}[theorem]{Lemma}
\newtheorem{remark}[theorem]{Remark}

\def\beq{\begin{equation}}
\def\eeq{\end{equation}}
\def\beqn{\begin{eqnarray}}
\def\eeqn{\end{eqnarray}}
\def\beqnn{\begin{eqnarray*}}
\def\eeqnn{\end{eqnarray*}}

\def\bq{{\bf q}}
\def\br{{\bf r}}

\def\bv{{\bf v}}

\def\bx{{\bf x}}
\def\by{{\bf y}}

\def\C{{\cal C }}

\newcommand{\norm}[1]{\left\| \, {#1} \, \right\|}
\newcommand{\abs}[1]{\left| {#1} \right| }
\newcommand{\parfirst}[2]{\frac{\partial {#1}}{\partial {#2}}}

\begin{document}

\title{{Curve Tracking Control for Legged
    Locomotion in Horizontal Plane}}
\author{ Fumin Zhang \\
 Department of Mechanical and Aerospace Engineering\\
         Princeton University\\
\{fzhang\}@princeton.edu } 
\maketitle
\begin{abstract}
We derive a hybrid feedback control law for the lateral leg spring (LLS) model 
so that the center of mass of a legged runner follows a curved path in 
horizontal plane.
The control law enables the runner to change the placement and the elasticity
of its legs to move in a desired direction. Stable motion 
along a curved path is
achieved using curvature, bearing and relative distance between the runner and
the curve as feedback.  Constraints on leg parameters determine the class of
curves that can be followed.
We also derive an optimal control law that stabilizes the orientation of
the runner's body relative to the velocity of the runner's center of mass. 
\end{abstract}

\section{Introduction}

 From tiny ants to large elephants, legged locomotion is the dominant
 method that animals use to move on the ground. Although the leg structures
 are vastly different across species, the mechanisms for walking,
 jumping or running obey strikingly similar principles. These
 similarities are captured by mathematical models such as  the spring-loaded inverted
pendulum (SLIP) model, see \cite{FullKod99}, and the lateral leg spring (LLS) model in
\cite{SchmittHolmes001, SchmittHolmes002,
  SchmittHolmes01,SchmittHolmes02}. 
 The LLS model describes
motion in the horizontal plane and the SLIP model describes motion
in the sagittal (vertical) plane. 

 We use the LLS model in this paper in designing  curve
 tracking control in the horizontal plane;
 the runner is modeled as a rigid body with two weightless
springs attached to a point in the body called the center of pressure
 (COP). Each spring represents legs on one side of the body.
 The COP is usually not coincident with the center of mass (COM). 
Legged locomotions can be
self-stabilized---the running or walking gaits stay close to
being periodic under disturbances---without feedback control. 
As indicated by a recent
review \cite{GAPK05}, the self-stabilized walking and running 
 happen when the runner moves along a straight line.

There are many interests in engineering practice to design and
build legged robots which are versatile on rough and uneven terrains.  Legged robots 
are greatly appreciated in applications such as searching and
rescuing missions  and
planet exploration. In most missions, the robots must be able to move along 
an arbitrary path. Feedback control is needed for tracking curved path as well as
stabilizing the periodic gaits.

In this paper, we develop feedback control law for the LLS model 
so that the COM follows a curved path. The legged locomotion modeled by the
LLS model is a hybrid system. Correspondingly, the tracking control contains a
discrete tracking algorithm which guarantees convergence to the desired curve and
a continuous law to control the leg parameters of the LLS model. We also 
develop an optimal control law to stabilize the posture of the rigid body.
Section \ref{sec:COM} serves as an introduction to the LLS model. The discrete tracking
algorithm is developed in Section \ref{sec:tr}. In Section \ref{sec:model}, control laws
are developed for the leg parameters to enable the discrete tracking algorithm.
The constraints on the parameters and the effects on the tracking behavior are discussed in
Section \ref{sec:constraint} and a modified continuous control law is introduced to handle the constraints. We then develop an optimal control law to achieve desired
posture for the rigid body in Section \ref{sec:rigid}. We provide simulation results in
Section \ref{sec:SIM}.

\section{\label{sec:COM}Motion of the COM}
\begin{figure}[htbp]
\centerline{
\psfig{figure=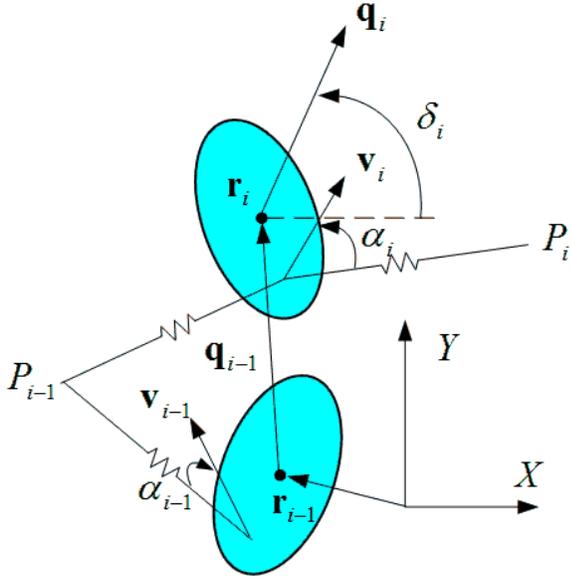 ,width=3in}
}
\caption{The LLS model for legged locomotion. One left stance followed
by one right stance are plotted. The center of mass (COM) is  translated
from $\br_{i-1}$ to $\br_i$. The velocities of the COM at the beginning
of each stance are the vectors
$\bv_{i-1}$ and $\bv_i$. The angle between $\bv_i$ and the spring (leg) is
$\alpha_i$. The angle between vector $\bq_i$ and the horizontal axis
$X$ is $\delta_i$. The positive directions for all angles are as shown. }
\label{fig:BugFigure1}
\end{figure}

 In the horizontal plane,  motion generated by the LLS model 
starts when the free end of one spring (or leg) is placed at a
touchdown point $P$. At this starting moment, the spring is at its free length
$\eta_0$. If the COM has a non-zero
initial velocity $\bv$  that is not
perpendicular to the spring, then the spring will be first compressed to 
a minimum length and later be restored  to its free length. 
This process, starting and ending with the spring at its free length
$\eta_0$, is called a {\it stance phase} or simply a {\it stance}. 
The COM moves from the starting
position to an ending position after a stance. Suppose that mechanical energy is
conserved during each stance, then the starting and ending speed of a stance is identical.
As shown in Figure \ref{fig:BugFigure1},
the end of one stance serves as the
beginning of the next stance with the touchdown point $P$
shifted from one side to the other. This allows us to distinguish left
stances from  right
stances based on which leg is supporting the body.

The rigid body moves forward as a result of switching between  left
and right stances. As shown in Figure \ref{fig:BugFigure1},
we use $\br_i$, $i=1,2,...,$
to denote the position of the COM in a lab fixed
coordinate frame at the beginning of the $i$th stance.
We use $\alpha_i$ to denote the angle between the velocity
$\bv_i$ and the spring at rest. 
 For a right stance, the angle is
measured counter-clockwise from the spring to the velocity vector. 
 For a left stance, the
angle is measured clockwise from the spring to the velocity
vector.   Under this convention for measuring angles, $\alpha_i$ 
 has to be within the interval $(0,\pi/2)$ to generate forward locomotion.

 We can view $\br_i$ as points on a curve $\Gamma$ which is
formed by 
straight line segments that connects
$\br_{i-1}$ with $\br_{i}$ for all $i$. This curve $\Gamma$ is not
the actual trajectory of the COM, but it intersects with the trajectory of the 
COM at the points $\br_i$.
We then let
$
     \bq_i = \br_{i+1}-\br_i
$ 
and define $q_i = \norm{\bq_i}$. We also define an angle $\delta_i$ as  the  angle between the
vector $\bq_i$ and the horizontal axis of the lab frame,  measured
counter-clockwise from the axis. This angle describes the direction of
curve $\Gamma$ for the $i$th stance. The motion of
the COM can now be described by a discrete system
\beq \label{equ:dsys}
    \br_{i+1} = \br_i + [q_i \cos \delta_i, q_i \sin \delta_i]^T.
\eeq
Next, we develop a boundary tracking algorithm for this discrete system.

\section{\label{sec:tr}Tracking a detected boundary}
\begin{figure}[htbp]
\centerline{
\psfig{figure=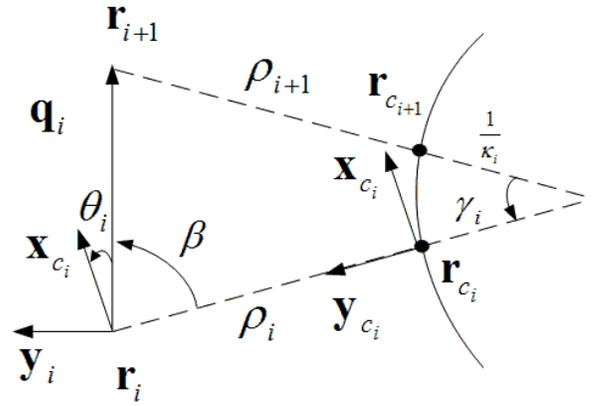 ,width=3in}
}
\caption{\it The movement of the COM near a boundary curve. $\theta_i$ is the angle between $\bq_i$ and $\bx_{{\rm c}_i}$. $\rho_i$ is the distance between the COM and the closest point. $\kappa_i$ is the curvature of the curve. $\gamma_i$ is the center angle of the arc connecting $\br_{{\rm c}_i}$ and $\br_{{\rm c}_{i+1}}$ }
\label{fig:BugFigure2}
\end{figure}

Suppose at the position $\br_i$,  the runner
is able to detect a segment of a boundary curve from sensor
information, c.f., \cite{ZhangPSKOcon04} and  \cite{lamperski05}. 
Suppose the runner is also
able to estimate a point on the boundary curve that has the minimum distance to the COM. We call this point the closest point $\br_{{\rm c}_i}$ shown in Figure 
\ref{fig:BugFigure2}. By selecting two extra points on the boundary
near $\br_{{\rm c}_i}$, the runner can
estimate the tangent vector to the boundary curve $\bx_{{\rm c}_i}$ and
the curvature of the boundary curve $\kappa_i$ using
algorithms summarized in \cite{ZhangPSKOcon04}. 
Here we suppose all estimates are perfect.

We let
$\rho_i$ represent the distance between the COM at $\br_i$ and the
closest point at $\br_{{\rm
  c}_i}$. We also let 
\beq
         \bx_i = \frac{\bq_i}{\norm{\bq_i}}
\eeq
be the unit vector in the direction of $\bq_i$. We can then
define two right handed frames---one at $\br_i$ and the other at
$\br_{{\rm c}_i}$---by defining two unit normal vectors $\by_i$ and
$\by_{{\rm c}_i}$ as shown in Figure \ref{fig:BugFigure2}. The angle $\theta_i$ between $\bx_i$ and $\bx_{{\rm c}_i}$
is defined by letting
\beq
        \cos\theta_i = \bx_i \cdot \bx_{{\rm c}_i} \;\; \mbox{and} \;\;
        \sin\theta_i=- \bx_i \cdot \by_{{\rm c}_i}.
\eeq

In Figure \ref{fig:BugFigure2}, the closest point on the
boundary moves from $\br_{{\rm c}_i}$ to $\br_{{\rm c}_{i+1}}$ during a
stance. 
We then assume that each stance  is short enough so that the curvature
of the boundary curve detected during this period can be approximated
by a finite constant $\kappa_i$. 
From Figure \ref{fig:BugFigure2}, we observe
\beq
       \beta= \frac{\pi}{2} - \theta_i\;.
\eeq
Therefore, according to the cosine law, we have
\beq \label{equ:dist}
       (\rho_{i+1}+\frac{1}{\kappa_i})^2 = (\rho_i+\frac{1}{\kappa_i})^2 - 2(\rho_i+\frac{1}{\kappa_i}) q_i \sin \theta_i + q_i^2.
\eeq
We can  view  $\rho_i$ as state variables and consider  
 $q_i$ as the control over step size  and $\theta_i$ as steering control.
Equation  (\ref{equ:dist}) describes the
controlled relative motion between the COM and the detected boundary curve.

Suppose that the step size $q_i$ has been determined. We need to
find $\theta_i$ such that $\rho_i$ converges to a desired value $\rho_{\rm c}$ as 
$i \to \infty$. 
If we can  find a feedback control law $f_i(\cdot)$ such that
\beq \label{equ:ri}
    \rho_{i+1} -\rho_i = f_i(\rho_i-\rho_{\rm c}),
\eeq
then we can choose the function $f_i(\cdot)$ so that $\rho_i-\rho_{\rm c} \to 0$. For example, we may let 
\beq
    f_i(\rho_i-\rho_{\rm c}) =-K_i\cdot (\rho_i-\rho_{\rm c})
\eeq
where $0<K_i<1$. Observe that 
\beq
      \rho_{i+1} -\rho_i = \rho_{i+1}-\rho_{\rm c} -(\rho_i -\rho_{\rm c}),
\eeq
 and this leads to
\beq
    \rho_{i+1}-\rho_{\rm c} = (1-K_i) (\rho_i-\rho_{\rm c}).
\eeq
Since $-1<1-K_i<1$, it is true that $\rho_i\to \rho_{\rm c}$ as $i \to \infty$.

We now define $\lambda_i=\rho_i+\frac{1}{\kappa_i}$. In order to achieve (\ref{equ:ri}),  we replace $(\rho_{i+1}+\frac{1}{\kappa_i})$ with $(\lambda_i+f_i)$ in (\ref{equ:dist}).  
From (\ref{equ:dist}), we can solve for
$\sin\theta_i$ as
\beqn \label{equ:control}
\sin \theta_i &=& \frac{-f_i^2-2\lambda_i f_i +q_i^2}{2 \lambda_i q_i} \cr
    &=& \frac{q_i}{2 \lambda_i}- \frac{f_i}{q_i}- \frac{f_i^2}{2\lambda_i q_i}\;.
\eeqn  
When $f_i=0$, the solution is $\theta_i = {\sin^{-1}}(q_i/2\lambda_i)$. This corresponds
to the runner running parallel to the boundary curve.

Note that when $\kappa_i\to 0$, then $\lambda_i \to \infty$.  The limit of equation (\ref{equ:control}) is
\beq \label{equ:straight}
\sin \theta_i = -\frac{f_i}{q_i} \;
\eeq  
which is equivalent to  $\rho_{i+1}-\rho_i=-q_i\sin\theta_i$. This equation indeed 
describes the relative motion between the COM and a straight line. Therefore, all
the results that will be obtained from (\ref{equ:control}) are applicable to (\ref{equ:straight}) by
letting $\lambda_i \to \infty$. 

\begin{lemma}
A solution exists for $\theta_i$ in equation (\ref{equ:control}) if and only if $f_i\in [-(2\lambda_i+q_i), \min\{-q_i,q_i-2 \lambda_i\}]$ or $f_i \in[\max\{-q_i, q_i-2\lambda_i\}, q_i]$. 
\end{lemma}
\begin{proof}
A solution exists for $\theta_i$ if and only if
\beq \label{equ:ineq1}
    \frac{-f_i^2-2\lambda_i f_i +q_i^2}{2 \lambda_i q_i}\le 1     
\eeq
and
\beq \label{equ:ineq2}
    \frac{-f_i^2-2\lambda_i f_i +q_i^2}{2 \lambda_i q_i}\ge -1.     
\eeq

From (\ref{equ:ineq1}) we have,
\beq
   f_i^2+2 \lambda_i f_i - (q_i^2-2\lambda_i q_i) \ge 0
\eeq
which is equivalent to
\beq
  (f_i+(2\lambda_i-q_i))(f_i+q_i) \ge 0.
\eeq
This inequality is satisfied if and only if
\beq \label{equ:con1}
          f_i \ge \max\{-q_i, q_i-2\lambda_i\}\mbox{ or } f_i \le \min\{-q_i, q_i-2\lambda_i\}.
\eeq

From (\ref{equ:ineq2}) we have,
\beq
   f_i^2+2 \lambda_i f_i - (q_i^2+2\lambda_i q_i) \le 0
\eeq
which is equivalent to
\beq
  (f_i+(2\lambda_i+q_i))(f_i-q_i) \le 0.
\eeq
This inequality is satisfied if and only if
\beq \label{equ:con2}
       -(2\lambda_i+q_i) \le  f_i \le q_i.
\eeq
Therefore, combining (\ref{equ:con1}) and (\ref{equ:con2}), we have found the 
necessary and sufficient condition for the existence of solution for $\theta_i$ as
\beq \label{equ:condition1}
   -(2\lambda_i+q_i)\le f_i \le \min\{-q_i,q_i-2 \lambda_i\} 
\eeq 
or
\beq \label{equ:condition2}
     \max\{-q_i, q_i-2\lambda_i\} \le f_i \le q_i. 
\eeq
\end{proof}

\begin{lemma} \label{pro:lm2}
 Suppose $0<q_i<2\lambda_i$. Let $f_i= -K_i (\rho_i-\rho_{\rm c})$. There exists $K_i\in (0,2)$ such 
that  (\ref{equ:condition2}) is satisfied and  $\theta_i$ in (\ref{equ:control}) has a solution.
\end{lemma}
\begin{proof}
If $\abs{\rho_i-\rho_{\rm c}}<\frac{1}{2}\min\{q_i,2\lambda_i-q_i\}$, we may let $K_i$ be any value in the interval $(0,2)$ and
 (\ref{equ:condition2}) is satisfied. If $\abs{\rho_i-\rho_{\rm c}}\ge \frac{1}{2}\min\{q_i,2\lambda_i-q_i\}$, we may let
\beq \label{eq:Kk}
 0<K_i <  \min\{2, \frac{q_i}{\abs{\rho_i-\rho_{\rm c}}},\frac{2\lambda_i-q_i}{\abs{\rho_i-\rho_{\rm c}}}\}
\eeq
which satisfies (\ref{equ:condition2}).
\end{proof}

The condition $q_i<2\lambda_i$ in Lemma \ref{pro:lm2} indicates that each step size
 should not be too big. If the curve has large curvature i.e. small turning radius, then the runner
must reduce its step size when being close to the curve. For a straight line, since $\lambda_i$ is
arbitrarily large, there is no constraints on the step size $q_i$.

\begin{theorem} \label{pro:t1}
Suppose the step size $q_i$ satisfy $0< q_i< 2\lambda_i$. Let $f_i= -K_i (\rho_i-\rho_{\rm c})$. Select $K_i\in (0,2)$ such that  $\theta_i$ in (\ref{equ:control}) has a solution. Then under the control $q_i$ and $\theta_i$, we have $\rho_i\to \rho_{\rm c}$ as $i\to \infty$.
\end{theorem}
\begin{proof}
Under the selection of $q_i$, $f_i$ and $K_i$, we have
\beq
       \rho_{i+1} -\rho_{\rm c} = (1-K_i) (\rho_i-\rho_{\rm c})
\eeq
for all $i$.
Since $-1<1-K_i<1$, the convergence is proved.
\end{proof}

Once $\theta_i$ is determined from (\ref{equ:control}), we can compute the direction of $\bq_i$ in the lab frame. This is
because
\beq \label{equ:angle1}
    \delta_i = \zeta_i - \theta_i
\eeq
where $\zeta_i$ is the angle between the horizontal axis of the lab frame and the vector $\bx_{c_i}$, 
measured counter clockwise from the horizontal axis. Knowing $q_i$ and $\delta_i$ allows us to
compute the position of the COM for the next stance from the discrete system given by equation (\ref{equ:dsys}).

\section{\label{sec:model} Control the LLS model}


In order to generate desired $q_i$ and  $\delta_i$ to control the
 COM movement,
the runner needs to change its leg placements or the elasticity of
the legs. We  investigate the dynamics during each stance to establish the relations
between the COM motion and the leg parameters.  

Leg parameters for a right stance and a left stance often differ only
by the sign. Since a right stance is always followed by a left stance and vice versa,
{\it we use the convention that the  stance $k$ is always a right stance and stance $k+1$ 
 is always a left stance}. Therefore, stance $k+2$ must be a right stance, etc. In
the following, we will only show detailed derivation for a right stance; similar
results for a left stance will be listed directly. 

 We set up a polar coordinate system at
 the touchdown point $P_k$ with the horizontal $\bx$-axis parallel to the
horizontal $X$-axis of the fixed lab frame.
Let $(\eta,\psi)$ be the polar coordinates of the COM in this
frame and let $\sigma$ describe the orientation of the  rigid
body. 
Then the
total energy is 
\beq \label{equ:energy}
        E = \frac{1}{2} m \dot \eta^2 + \frac{1}{2} m \eta^2 \dot \psi^2 +
        \frac{1}{2} I \dot \sigma^2
+ V(\eta)
\eeq
 where we assume the spring has potential energy $V$ which depends
 only on its length. 


During each stance, the dynamics can be described as a continuous nonlinear Hamiltonian system; the Hamilton equations are developed in \cite{SchmittHolmes001}. 
 The system is not integrable when the distance between
the COM and the COP is nonzero. 
In this case numerical methods are necessary to compute the trajectory of
the COM from knowledge of the  states $(\eta,\psi,\sigma)$. 

To illustrate analytical insights for the tracking problem, it is much easier
to study the case when the COP and COM coincide. This is because the corresponding system is integrable. The Hamilton equations for the
dynamics of the COM are
\beqn \label{equ:LLS}
      p_\eta &=& m \dot \eta \cr
      p_\psi &=& m \eta^2 \dot \psi \cr
      \dot p_\eta &=& \frac{1}{m \eta^3} p_\eta^2 -
      \parfirst{V}{\eta} \cr
\noalign{\smallskip}
      \dot p_\psi &=& 0.
\eeqn  
\begin{figure}[htbp]
\centerline{
\psfig{figure=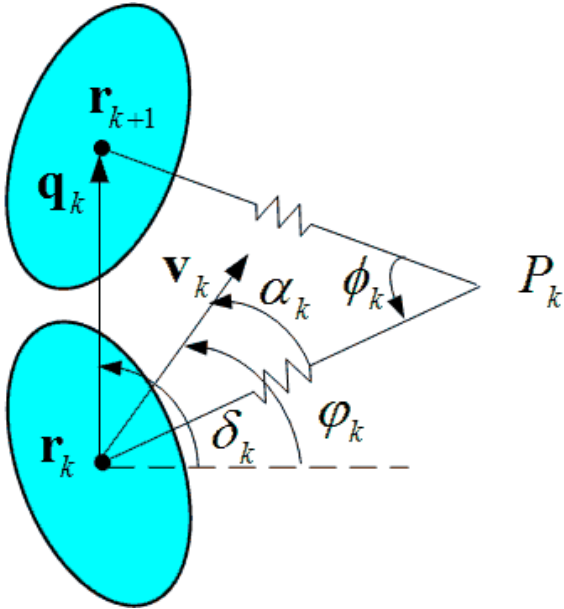 ,width=3in}
}
\caption{\it The LLS model with COM coincide with COP. Angle $\phi_k$ is
the center angle swept during the stance. Angle $\alpha_k$ is the angle between the leg and the velocity $\bv_k$ at the moment of touch down.}
\label{fig:BugFigure3}
\end{figure}

 We plot one right stance in Figure \ref{fig:BugFigure3}.
Let $\bv_k$ be the velocity of the COM at the beginning of the $k$th stance. 
 This $\bv_k$ provides initial value for the states $p_\eta$ and $p_\psi$ of the system (\ref{equ:LLS}).
Because
mechanical energy is conserved, the speed $v_k=\norm{\bv_k}$ satisfies $v_{k+1}=v_{k}$. We let $\alpha_k$ represent the angle measured from the leg to the velocity $\bv_k$ at the moment of touchdown, and we call $\alpha_k$ {\it the leg placement parameter}. We  use $\varphi_k$ to measure
 the direction of velocity in the lab frame.
By using simple geometric relationships in Figure
\ref{fig:BugFigure3}, we have
\beq
         \frac{\pi}{2}-\frac{\phi_k}{2}-\alpha_k=  \delta_k -\varphi_k.
\eeq
Therefore, using equation (\ref{equ:angle1}),
\beq \label{equ:alpha}
     \zeta_k -\varphi_k-\theta_k = \frac{\pi}{2}-\alpha_k-\frac{\phi_k}{2}.
\eeq
Note that $(\zeta_k-\varphi_k)$ 
is the relative angle between $\bv_k$ and $\bx_{c_k}$: the  tangent vector to the boundary curve at the
closest point. We know $\theta_k$ is the angle between vector 
$\bq_k$ and the same tangent vector  $\bx_{c_k}$. Therefore, $(\zeta_k -\varphi_k-\theta_k)$ is the angle between $\bv_k$ and $\bq_k$. From (\ref{equ:alpha}), we see that changing $\alpha_k$ will change the direction of the COM movement  $\bq_k$. This is also true for a left stance where we have
\beq \label{equ:alpha2}
     \zeta_{k+1}- \varphi_{k+1}-\theta_{k+1}= - (\frac{\pi}{2}-\alpha_{k+1}-\frac{\phi_{k+1}}{2}).
\eeq 

Another relationship we can derive from Figure \ref{fig:BugFigure3} is
\beq \label{equ:qphi}
    q_k = 2 \eta_k \sin \frac{\phi_k}{2}
\eeq
where $\eta_k$ represents the leg length at the moment of touchdown.

In (\ref{equ:qphi}) and
(\ref{equ:alpha}), $q_k$ is the distance the runner wants to
travel in one stance, the angle $\theta_k$ can be solved from (\ref{equ:control}),
and the angle $(\zeta_k-\varphi_k)$ is known. 
We want to solve for the leg parameters $\alpha_k$ and  $\eta_k$, but $\phi_k$ is still unknown. This unknown 
can be solved from the continuous system equations (\ref{equ:LLS}).

At the starting position of the $k$th stance $\br_k$ and the ending position of the $k$th stance $\br_{k+1}$, by conservation of the angular
momentum, we
have
\beq
        p_{\eta_{k+1}}= p_{\eta_{k}} = \eta_k^2 \dot \psi = \eta_k v_k \sin\alpha_k.
\eeq
Using the method of integration by quadrature, 
c.f. \cite{Arnold89}, we can compute
 the center swing angle
$\phi_k$  for each stance
as
\beq \label{equ:constraints}
        \phi_k = 2 \int_{\eta_k}^{\eta_{\rm min}}
       \frac{\frac{p_{\eta_k}}{\eta^2}}{\pm\sqrt{2E-\frac{p^2_{\eta_k}}{\eta^2}-2V(\eta)}} d \eta
\eeq
where $\eta_{\rm min}$ is the shortest length of the spring during the stance. When $\eta=\eta_{\rm min}$, we have $\dot
\eta=0$. Thus we can solve $\eta_{\rm min}$ from 
\beq
         2 E - \frac{p^2_{\eta_k}}{\eta^2_{\rm min}}- 2V(\eta_{\rm min})=0\;.
\eeq
  Explicit formulas can be derived for $\phi_k$ when we use
the linear spring potential $V= b_k (\eta-\eta_k)^2$ where $b_k$ is the spring constant for the $k$th stance. These formulas require
the use of elliptic functions
\cite{SchmittHolmes001}.     

Since $\phi_k$ is now a known function of $\alpha_k$, $\eta_k$, and $b_k$, we can solve for any two of $\alpha_k$, $\eta_k$,
 and $b_k$ from (\ref{equ:alpha}) and (\ref{equ:qphi}) when keeping the other parameter  constant.
For a runner, controlling $\alpha_k$ means to find the appropriate angle
between its leg and the direction of the COM motion.
On the other hand,
 as
reported by Jindrich and Full in \cite{JindrichFull99},  the cockroaches control the 
length $\eta_k$ by stretching or compressing their legs when
turning. 
We see that changing $\eta_k$ will affect both $p_{\psi_k}$ and $V(\eta)$. This changes
$\phi_k$ and hence controls $(\zeta_k -\varphi_k-\theta_k)$.
Another means of steering is to
change the potential energy $V(\eta)$ ,e.g., change the spring constant $b_k$, which also controls $\phi_k$.

If the conditions in Theorem \ref{pro:t1} are satisfied, equation (\ref{equ:control}) always has a
solution for $\theta_i$. 
 The equations (\ref{equ:alpha}), (\ref{equ:qphi})
 and (\ref{equ:constraints}) can be solved to implement the control $\theta_i$.  We call this
method {\it the inverse  method}.   
Note that finding solutions for $\alpha_k$ and $\eta_k$ often requires numerical methods
because $\phi_k$ is not a simple function of $\alpha_k$ and $\eta_k$.

\section{\label{sec:constraint}Tracking Behavior Under Constraints}
For every stance, the LLS model generates the COM movement $\bq_k$ by controlling  parameters such as 
$\alpha_k$, $\eta_{k}$ and $b_k$. In practice, these parameters all have to be bounded. These bounds
post constraints on the possible $\bq_k$ that can be generated by the LLS model.
In this section, we first discuss the constrained COM movement and  investigate
 the constrained tracking behavior  when a runner is running along a curve path. We then derive a
new control law with proved convergence under the constraints.

\subsection{the constraints}
\begin{figure}[htbp]
\centerline{
\psfig{figure=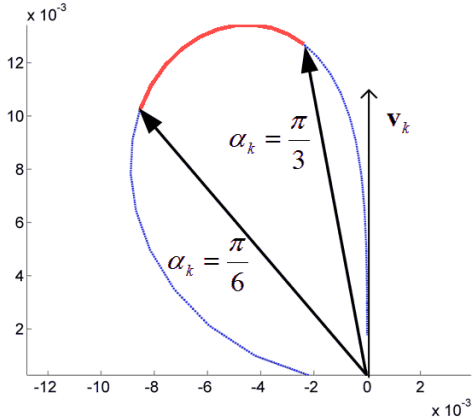 ,width=3in}
}
\caption{The possible $\bq_k$ generated by a right stance for an LLS model. $\bv_k$ is the velocity
of the COM. The units are in meters. The curve (dotted and solid) illustrates the
end points for vector $\bq_k$ starting from the origin when $\alpha_k$ changes from $0$ to $\pi/2$ while other
parameters are constant. The solid segment corresponds to $\alpha_k \in [\pi/6, \pi/3]$.}
\label{fig:BugFigure4}
\end{figure}

\begin{figure}[htbp]
\centerline{
\psfig{figure=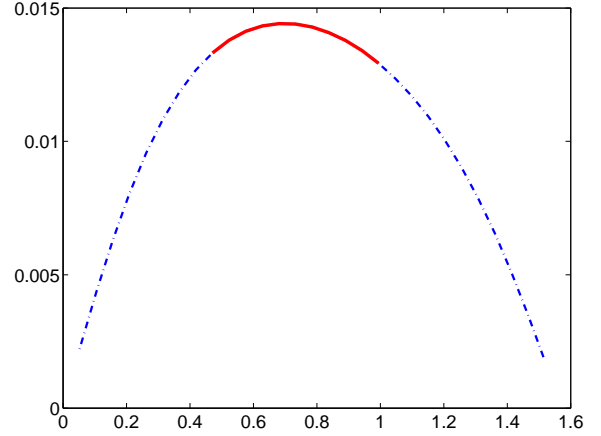 ,width=3in}
}
\caption{The length  $q_k$ (in meters) traveled by the COM during a right stance for an LLS model as a function of
leg placement angle $\alpha_k$ (in radians). The solid segment corresponds to $\alpha_k \in [\pi/6, \pi/3]$.}
\label{fig:BugFigure5}
\end{figure}

To generate forward locomotion, the relative angle $\alpha_k$ between the leg and the COM velocity $\bv_k$ should be bounded within the interval  $(0, \pi/2)$. Figure \ref{fig:BugFigure4} illustrates the possible $\bq_k$ that can be produced by changing $\alpha_k$ for a right stance when $\eta_k$ and $b_k$ are held constant. The parameters for the plotted LLS model are $m=2.5 {\rm g}$, $v_k=0.2 {\rm m/s}$, $\eta_k = 1.7 {\rm cm}$ and $b_k=1.05 {\rm N/m}$ which are typical for a cockroach. When $\alpha_k=0$ and $\alpha_k=\pi/2$, we have $q_k=0$. Therefore, in order to move forward effectively, the angle $\alpha_k$ must be within an interval $[\alpha_{\rm min}, \alpha_{\rm max}]$ with $\alpha_{\rm min}>0$ and $\alpha_{\rm max}< \pi/2$. The solid segment in Figure \ref{fig:BugFigure4} illustrates the possible $\bq_k$ between $\alpha_{\rm min}=\pi/6$ and $\alpha_{\rm max}=\pi/3$. We also plot the length $q_k$  as a function of the leg placement angle $\alpha_k$ in Figure \ref{fig:BugFigure5}. There the maximum $q_k$ is $1.44{\rm cm}$. When $\alpha_k$ is within $[\pi/6, \pi/3]$, the minimum $q_k$ is $1.24{\rm cm}$. The changes in $q_k$ is not big for a wide range of $\alpha_k$. This is typical for  LLS models.

\begin{figure}[htbp]
\centerline{\psfig{figure=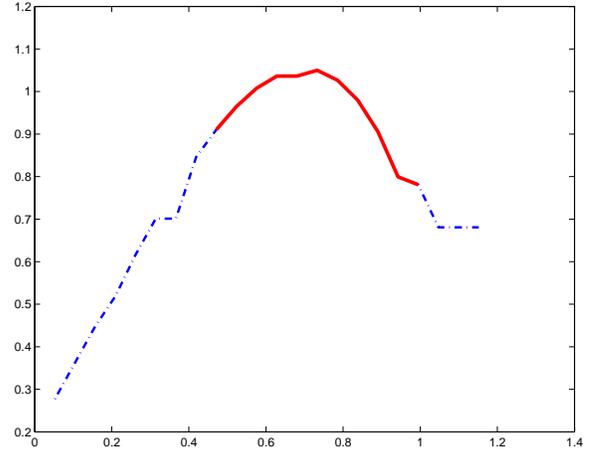 ,width=3in}}
\caption{The spring constant  $b_k$ (in N/m) as a function of
leg placement angle $\alpha_k$ (in radians) to keep $q_k=1.44\,{\rm cm}$. The solid segment corresponds to $\alpha_k \in [\pi/6, \pi/3]$.}
\label{fig:BugFigure6}
\end{figure}

 The above example suggests that it is possible to keep $\eta_k$, $q_k$ and $\phi_k$ constant
for each stance. We control the spring constant $b_k$ and the leg placement angle $\alpha_k$. The advantage of this strategy is that the distance traveled by the COM is identical for
every stance. This fact can help us analyze the tracking behavior later. For the LLS model plotted in Figure \ref{fig:BugFigure4} and \ref{fig:BugFigure5}, in order to keep $q_k=1.44{\rm cm}$, we plot the spring constant $b_k$ as a function of
the leg placement angle $\alpha_k$ in Figure \ref{fig:BugFigure6}. For $\alpha_k \in [\pi/6, \pi/3]$, $b_k$ lies between $0.78\,{\rm N/m}$ and $1.06\,{\rm N/m}$. This range is not difficult to implement.      

\begin{figure}[htbp]
\centerline{
\psfig{figure=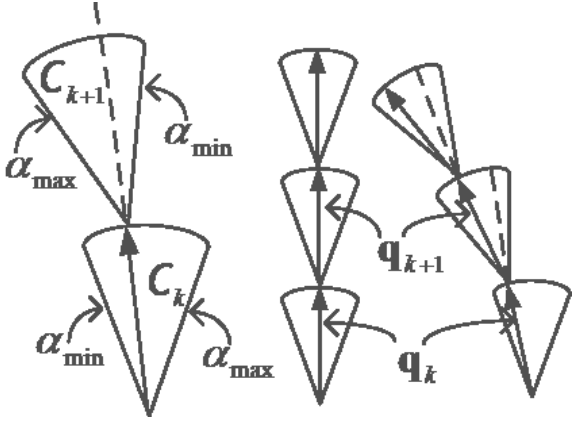 ,width=3in}
}
\caption{ The cones flip from one side to another when running. On the left, the cones $\C_k$ and $\C_{k+1}$ are symmetric with respect to
$\bq_k$ which is the solid arrow. In the middle, the runner is running along a straight line. On the right, the runner is running along a curve path in the counter
clockwise direction.}
\label{fig:BugFigure7}
\end{figure}

With $q_k$  constant
for each stance, the possible movement of the COM can be depicted by a cone $\C_k$. The two edges of the
cone correspond to $\alpha_k = \alpha_{\rm min}$ and $\alpha_k = \alpha_{\rm max}$. The length of both 
edges are $q_k$. Figure \ref{fig:BugFigure7} illustrates
cone $\C_{k+1}$ and cone $\C_k$. $\C_{k+1}$ grows from the end point of $\bq_k$ which lies in the circular 
arc of $\C_k$. We found that $\C_{k+1}$ is the mirror image of  $\C_{k}$ with $\bq_k$ being the axis of symmetry.
This is because the velocity vector $\bv_{k+1}$ and $\bv_k$ are symmetric with respect to $\bq_k$, which can
be proved by solving equations (\ref{equ:LLS}). Therefore, as the runner moves forward, the cone will flip
from side to side.

\subsection{running along a curve path and robustness}
We use the index $i$ for all right and left stances.
When running parallel to a desired curved path, the COM movement satisfies
$\rho_i=\rho_c$ for all $i$. Therefore, we have $f_i=0$ in equation
 (\ref{equ:control}). The following conditions are necessary:
\begin{itemize}
\item[A1)]{$\bq_i \in \C_i$; }
\item[A2)]{$\theta_i = \sin^{-1}(q_i/2\lambda_i)$.}
\end{itemize} 
Condition A1 requires that the COM movement belongs to the cone that is feasible for the
constrained model. Condition A2  requires that $\bq_i$ is parallel to the desired curve.

\begin{figure}[htbp]
\centerline{
\psfig{figure=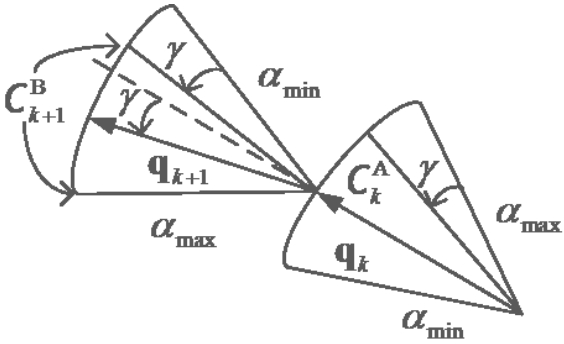,width=3in}
}
\caption{The sub-cones $\C^{\rm A}_k$ and $\C^{\rm B}_{k+1}$. $\bq_k$ is the middle line of $\C^{\rm A}_k$, and $\bq_{k+1}$
is the middle line of $\C^{\rm A}_k$. The angle difference between $\bq_{k+1}$ and $\bq_k$ is $\gamma$.   }
\label{fig:BugFigure8}
\end{figure}

If the desired curve is a straight line segment, then $\theta_i=0$ in condition A2. As the runner moving forward, 
the cone $\C_i$ will be flipping side to side with respect to the straight line. In this case we have
$\alpha_i=\alpha_{i+1}$. The robustness of
this behavior is determined by the size of the cone and the value of $\alpha_i$. If
 we choose the value of $\alpha_i$ so that the COM movement $\bq_k$ is always in
the middle of the cone, then the tracking behavior is the most robust. See the middle
figure in Figure \ref{fig:BugFigure7}.

 If the desired curve is convex with positive curvature, then $\theta_i\ne 0$.
To find out $\theta_i$, we study the $k$th and $(k+1)$th stance, i.e., a right stance
followed by a left stance.
 For convenience we let
\beq
 \gamma_k =2 \sin^{-1}\frac{q_k}{2\lambda_k}. 
\eeq
From equation (\ref{equ:theta2}) in Appendix \ref{apxa} and using the fact that
 $\phi_k=\phi_{k+1}$, we know that $\theta_k$ and $\theta_{k+1}$ satisfies
\beq \label{equ:theta3}
  \theta_{k+1} -\theta_k = -(\alpha_{k+1} -\alpha_k) +  \gamma_k.
\eeq 
 This relation can also be observed from Figure \ref{fig:BugFigure2}. 
Condition A2 implies that $\theta_{k+1}=\theta_k= \gamma_k/2$. Therefore, when running
along  a convex curve, $\alpha_{k+1} -\alpha_k =  \gamma_k$ should be satisfied.
We have similar relation  for stance $k+1$ and $k+2$, i.e., a left stance followed by a
right stance:  $\alpha_{k+2} -\alpha_{k+1} = - \gamma_{k+1}$. We can then write
$\alpha_{i+1} -\alpha_{i} = \pm \gamma_i$ for all stances.
This equation requires that $\gamma_i$ must be less than $(\alpha_{\rm max}-\alpha_{\rm min})$. This
implies that $\lambda_i$, the instantaneous radius of the curve must satisfy
\beq
      \lambda_i>  \frac{q_i}{2\sin\frac{\alpha_{\rm max}-\alpha_{\rm min}}{2}}.
\eeq
This condition is stricter than $\lambda_i>q_i/2$ required by Lemma \ref{pro:lm2}.
Hence the constraint on $\alpha_i$ puts a tighter restriction on the curvature of the curve that
can be traced.

 We divide $\C_i$ into two sub-cones. Let 
$\C^{\rm A}_i$ be the cone for $\alpha \in [\alpha_{\rm min},\alpha_{\rm max}- \gamma_i]$. Let
$\C^{\rm B}_i$ be the cone for $\alpha \in [\alpha_{\rm min}+ \gamma_i,\alpha_{\rm max}]$.
Because $\C_i$ flips and $\alpha_{i+1} -\alpha_i = \pm \gamma_i$, if $\bq_i$ belongs to $\C_i^A$, then
$\bq_{i+1}$ belongs to $\C_{i+1}^{\rm B}$.
Now consider a right stance $k$ followed by a left stance $(k+1)$. 
When running along
a convex curve in the counter clockwise (CCW) direction, the runner must have $\alpha_k \in \C^{\rm A}_k$  
and $\alpha_{k+1} \in \C^{\rm B}_{k+1}$, see  Figure \ref{fig:BugFigure8}.
When running in the  clockwise (CW) direction, the runner must have
$ \alpha_k \in \C^{\rm B}_k$ and $\alpha_{k+1} \in \C^{\rm A}_{k+1}$. 

To increase the robustness of the tracking behavior, we should let the COM movement $\bq_i$
be close to the middle of $\C_i$. In the case of convex curves, the best choice is to let $\bq_i$ and $\bq_{i+1}$ be symmetric
with respect to the middle line of $\C_i$.
The middle lines of $\C^{\rm A}_i$ and $\C^{\rm B}_i$ 
are symmetric with respect to the middle line of $\C_i$. The angle between the middle lines of  $\C^{\rm A}_i$ and $\C^{\rm B}_i$ is $\gamma_i$. If the curve has constant curvature, then $ \gamma_i$ is constant for all $i$. This implies that the middle lines of $\C^{\rm A}_i$ and $\C^{\rm B}_{i+1}$ are symmetric with respect to the middle line of $\C_i$.
Therefore, we may choose $\bq_i$ to be the middle line of either  $\C^{\rm A}_i$ or  $\C^{\rm B}_i$ for maximum robustness.
If the curve has a changing positive curvature, then $\gamma_i$ can be different from stance to stance. This ``middle line'' strategy can not be enforced. In this case one can choose $\bq_i$ to be as close to the middle lines as possible.

If the desired curve is not convex, 
the tracking behavior can be viewed as switching between tracking a locally convex curve
in the CCW direction and  in the CW direction. The switching depends on how the curve changes from locally
convex to locally concave. No general conclusions can be drawn regarding which part of the cone $\C_i$ is used for
a stance. In this case the tracking behavior is not a ``steady state'' . 

\subsection {\label{sec:slide}The approximation method}

The inverse method fails when constrained solutions from (\ref{equ:alpha}), (\ref{equ:qphi})
 and (\ref{equ:constraints}) do not exist. We design a  new control law that is able to find leg parameters that satisfies all constraints. The control law also guarantees convergence to the desired curve from generic initial 
conditions. We call this method the {\it approximation method}. 

We investigate the LLS model satisfying $q_i=q$ for all $i$ where 
$q$ is a positive constant. To keep $q_i=q$, the runner controls both the leg placement
angle $\alpha_i$ and the spring parameter  $b_i$. 
We assume that the desired curve path is either a straight line or a circle i.e.
$\kappa_i$ is constantly $\kappa$ for all $i$. This assumption can be relaxed to convex curves with slowly varying curvature.
We further assume that the step size $q$ is much smaller than the radius of the curve path. Under these assumptions, $\rho_{i+1}-\rho_i\ll 2(\rho_i+\frac{1}{\kappa})$, and equation (\ref{equ:control})
can be simplified to
\beq \label{equ:simsys}
\rho_{i+1}- \rho_{\rm c} = \rho_i-\rho_{\rm c} + \frac{q^2}{2\lambda_i}- q \sin \theta_i
\eeq
where $\lambda_i=(\rho_i+\frac{1}{\kappa})$. 
We then view $\sin\theta_i$ as the second state variable other than $\rho_i-\rho_{\rm c}$.
It satisfies
\beq \label{equ:simtheta}
       \sin \theta_i = \sin\left(\zeta_i -\varphi_i \mp (\frac{\pi}{2}-\alpha_i-\frac{\phi_i}{2})\right).
\eeq
when all the angles are mapped to the interval $(-\pi/2,\pi/2)$. We can apply the backstepping technique for discrete systems c.f. \cite{ordonezpassino03} to
the system described by (\ref{equ:simsys}) and (\ref{equ:simtheta}).

Let $f_i=-K_i(\rho_i-\rho_{\rm c})$.
According to Theorem \ref{pro:t1}, we can select $K_i$ such that (\ref{equ:control}) can be
solved for $\theta_i$. We let this solution be $\tilde \theta_i$ i.e.
\beqn
     \sin \tilde \theta_i &=& \frac{-f_i^2-2\lambda_i f_i +q^2}{2 \lambda_i q} \cr
    &=& \frac{q}{2 \lambda_i}- \frac{f_i}{q}- \frac{f_i^2}{2\lambda_i q}.
\eeqn
Using the fact that 
\beq
f_i^2=(\rho_{i+1}-\rho_i)^2\ll 2(\rho_i+\frac{1}{\kappa})= 2\lambda_i,
\eeq
we have
\beq \label{theta4}
      q \sin \tilde \theta_i = \frac{q^2}{2 \lambda_i}- f_i.
\eeq 
Note that $\tilde \theta_i$ is
different from the state variable $\theta_i$ in (\ref{equ:simsys}) and (\ref{equ:simtheta}). Here $\tilde \theta_i$
is a function of $\rho_i$ and can be viewed as the desired value for the state $\theta_i$.   

We can then solve for the term $\frac{q^2}{2 \lambda_i}$ from (\ref{theta4}) and substitute this term in the right hand side of (\ref{equ:simsys}). This yields
\beq \label{theta5}
     \rho_{i+1}- \rho_{\rm c} = \rho_i-\rho_{\rm c} + f_i - q( \sin \theta_i-\sin \tilde \theta_i).
\eeq
We define 
\beq
       \tilde \rho_{i+1} \equiv \rho_i-\rho_{\rm c} + f_i= (1-K_i) (\rho_i-\rho_{\rm c}).
\eeq
This $\tilde \rho_{i+1}$ can be viewed as the desired value for the state $\rho_{i+1}-\rho_{\rm c}$.
With the help of $\tilde \theta_i$ and $\tilde \rho_{i+1}$, we rewrite  (\ref{theta5}) as
\beq \label{equ:sthk3}
    \rho_{i+1}- \rho_{\rm c}-\tilde \rho_{i+1} = -q (\sin \theta_{i}-\sin\tilde \theta_i).
\eeq

 We design $\alpha_{i}$ as a feedback law so that  the right hand side of (\ref{equ:simtheta}) satisfies
\beq \label{equ:slcontrol}
      \sin\left(\zeta_i -\varphi_i \mp (\frac{\pi}{2}-\alpha_i-\frac{\phi_i}{2}) \right) = \sin \tilde \theta_{i} + \frac{\tilde K_i}{q}(\rho_i-\rho_{\rm c}- \tilde \rho_{i})
\eeq
where $\tilde K_i$ is a scalar which will be determined later.
Thus  (\ref{equ:simtheta}) becomes
\beq \label{equ:simthk2}
    \sin \theta_{i} -\sin \tilde \theta_{i} = \frac{\tilde K_i}{q}(\rho_i-\rho_{\rm c}- \tilde \rho_{i}).
\eeq
We show that the closed loop system given by (\ref{equ:sthk3}) and (\ref{equ:simthk2}) converges to the state where
 $\theta_i=\tilde \theta_i$ and $\rho_i=\rho_{\rm c}$.

\begin{lemma} \label{pro:Tm2}
Consider the system given by (\ref{equ:simsys}) and (\ref{equ:simtheta}).
 Let $K_i$ and $\tilde K_i$ be such that $|1-K_i|<1$ and $|\tilde K_i| < 1$. Suppose (\ref{equ:slcontrol}) has a solution
for $\alpha_{i}$. Then as $i\to \infty$, we must have $\theta_i \to \tilde \theta_i$ 
and $\rho_i\to \rho_c$.  
\end{lemma}
\begin{proof}
If (\ref{equ:slcontrol}) has a solution for $\alpha_{i}$, then (\ref{equ:sthk3}) and
(\ref{equ:simthk2}) hold. Therefore
\beq
   \sin \theta_{i}-\sin \tilde \theta_{i} =-\tilde K_i ( \sin \theta_{i-1}-\sin \tilde \theta_{i-1}).
\eeq
Since $|\tilde K_i| < 1$,  it is true that  $\sin \theta_{i}-\sin \tilde \theta_{i}\to 0$ as $i \to \infty$.
Meanwhile, from (\ref{equ:sthk3}), we conclude $\rho_{i+1}-\rho_{\rm c}-\tilde \rho_{i+1}\to 0$ as $i\to \infty$. This implies that $\rho_{i+1}-\rho_{\rm c}\to (1-K_i)(\rho_i-\rho_{\rm c})$. Since $|1-K_i|<1$, we conclude $\rho_i\to \rho_{\rm c} $ as $i\to \infty$.
\end{proof}

 If we allow $|\tilde K_i|$ to be arbitrary large then (\ref{equ:slcontrol})
always has solutions for $\alpha_{i}$. By selecting proper value for $\tilde K_i$, 
we can find a solution that
satisfies the constraints for $\alpha_{i}$.  However, Lemma \ref{pro:Tm2} requires that
$|\tilde K_i|<1$ to achieve asymptotic convergence. We want to find out when (\ref{equ:slcontrol}) fails to have a solution for $\alpha_{i}$ if $|\tilde K_i| <1$.

If we can find $\alpha_{i}\in [\alpha_{\rm min},\alpha_{\rm max}]$ such that $\sin\theta_{i}-\sin\tilde \theta_{i}=0$. Then we can let $\tilde K_i=0$ and a solution for $\alpha_i$ exists. This is exactly the inverse
method. 

If we can not find a constrained $\alpha_{i}$ such that $\sin\theta_{i}-\sin\tilde \theta_i=0$,
then let
\beq \label{equ:Mi}
\tilde M_i =\min_{\alpha_{i}\in[\alpha_{\rm min},\alpha_{\rm max}]}\{|\sin\theta_{i}-\sin\tilde \theta_{i}| \}.
\eeq
 We  let
\beq \label{equ:tki}
 \tilde K_i = \frac{\tilde M_i q}{\rho_i-\rho_{\rm c}- \tilde \rho_{i}}.
\eeq
If
\beq \label{equ:rhok}
 |\rho_i-\rho_{\rm c}- \tilde \rho_{i}| > \tilde M_i q
\eeq 
is true, then $|\tilde K_i|<1$.
 Equation (\ref{equ:slcontrol}) becomes
$\sin\theta_{i} - \sin \tilde \theta_{i} = \tilde M_i$, and a 
solution for $\alpha_{i}$ can be found.  
Therefore, when a solution for $\alpha_{i}$ can
not be found for $\tilde K_i$ given in (\ref{equ:tki}), we must have
\beq \label{equ:rhok2}
 |\rho_i-\rho_{\rm c}- \tilde \rho_{i}| \le \tilde M_i q.
\eeq
We argue that this implies that the distance between the runner and the desired path becomes sufficiently small as $i\to \infty$.

We define 
$\tilde M = \sup_i \tilde M_i$.  
Since $|\sin\theta_{i}-\sin\tilde \theta_{i}|\le 2$ regardless of the value of $\theta_{i}$ and
$\tilde \theta_{i}$, it is true that $\tilde M\le 2$.
The following theorem claims that the controlled movement of the COM converges to a small neighborhood of the desired curve.
\begin{theorem} \label{pro:cor1}
Consider the system given by (\ref{equ:simsys}) and (\ref{equ:simtheta}) controlled by the approximation
 method. Suppose the gain $K_i$ is constantly $K$ for all $i$ and
$|1-K|<1$. We determine the value for leg placement angle $\alpha_{i}$
by solving (\ref{equ:slcontrol}). For every $i$, we let $\tilde K_i$ be given by (\ref{equ:tki}).
Then we must have  $\lim_{i\to \infty}|\rho_i- \rho_c|\le\tilde M q/(1-|1-K|)$.  
\end{theorem}

\begin{proof}
 If  $|\tilde K_i| < 1$ for all $i$, Lemma \ref{pro:Tm2} claims that the system converges to $\rho_i=\rho_{\rm c}$.
Then  $\lim_{i\to \infty}|\rho_i- \rho_c|=0<\tilde M q/(1-|1-K|)$.

If for some time indices $j$, $|\tilde K_j|\ge 1$,  then condition (\ref{equ:rhok2}) must be true. We must have 
\beq
        |\rho_j-\rho_{\rm c}- \tilde \rho_{j}| \le \tilde M_j q \le \tilde M q
\eeq
This implies that 
\beq \label{equ:tt1}
|\rho_j-\rho_{\rm c}| \le |1-K|\cdot|\rho_{j-1}-\rho_{\rm c}| +\tilde M q.
\eeq
For the rest of time indices $i$ where $|\tilde K_i|<1$,  equation (\ref{equ:sthk3})
implies that
\beq
 |\rho_i-\rho_{\rm c}| \le |1-K|\cdot|\rho_{i-1}-\rho_{\rm c}| +|\sin\theta_{i-1}-\sin\tilde \theta_{i-1}| q
\eeq
Since $\tilde K_{i-1}$ is given by (\ref{equ:tki}),  
 We must have $|\sin\theta_{i-1}-\sin\tilde \theta_{i-1}|=\tilde M_{i-1}$. This  implies that (\ref{equ:tt1}) is satisfied for 
indices $i$. Therefore (\ref{equ:tt1}) is satisfied for all $i$ and $j$.
Because $|1-K|<1$,  we conclude that $\lim_{i\to \infty}|\rho_i-\rho_{\rm c}|\le \tilde M q/(1-|1-K|)$.
\end{proof}

\begin{remark}
From Theorem \ref{pro:cor1}, we conclude that larger step size (bigger $q$) and faster convergence rate in relative
distance (bigger $K$) may cause larger 
tracking error. This agrees with intuition.
The Theorem is conservative because we can make $K_i$ adaptive. When $|\rho_i-\rho_{\rm c}|$
small, we can let $K_i$ to be close to $1$ to reduce  the possible tracking error. Also, $\tilde M$ does
not have to be the supremum of $\tilde M_i$ for all $i$. Instead, we can let $i$ be larger than any
finite time. This  may reduce $\tilde M$.
Another observation is that by increasing $\alpha_{\max}-\alpha_{\min}$, we can reduce $\tilde M$, hence
reduce the tracking error. This also agrees with intuition.
\end{remark}

\begin{remark}
Equation (\ref{equ:Mi}) is used to compute $\alpha_i$ that minimizes the difference 
between $\theta_i$ and $\tilde \theta_i$. This is why the method is called the approximation method. 
Such a method can be devised heuristically without applying the backstepping technique. But the backstepping technique helps to justify convergence. One can also derive other convergent method which have different performance than the current one. But from the proof of Lemma \ref{pro:Tm2} we conclude that the approximation method gives the fastest convergence rate among all such methods derived from the backstepping procedure.   
\end{remark}

\section{\label{sec:rigid}Controlled Rigid Body Dynamics}
Unlike stable running along a straight line, the rigid body angular
momentum  $p_\sigma$
should not be zero for stable running along a curve; otherwise no
turning can happen. Since our goal is to control the runner to 
the desired curve, even if the final stable running is along a straight
line, the runner need to turn in order to move to that line from
an arbitrary initial position. 
The LLS model dynamics for the rigid body is
\beqn \label{equ:sigmacon}
  p_\sigma &=& I \dot \sigma \cr
      \dot p_\sigma &=& \tau
\eeqn
where $\tau$ represents the torque to produce turning 
for the rigid body. Runners can produce this
torque by using muscles connecting legs and body or by 
changing force distribution over multiple legs. 
\begin{figure}[htbp]
\centerline{
\psfig{figure=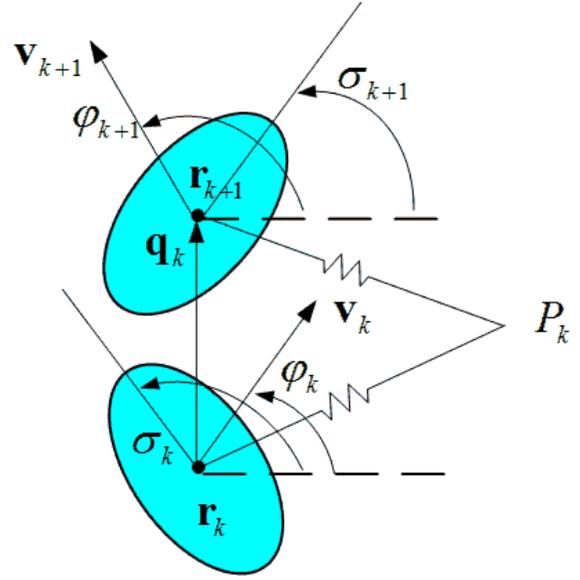,width=3in}
}
\caption{The posture of the rigid body at the beginning and the end of a stance.  }
\label{fig:BugFigure9}
\end{figure}

As indicated in Figure \ref{fig:BugFigure9},
 the angle between the
body axis and the velocity of the COM is $(\sigma_k-\varphi_k)$.
For stable running, we want to achieve the following gaits: 
after the $k$th stance, the ending angle $(\sigma_{k+1}-\varphi_{k+1})$ differs 
from the starting  $(\sigma_k-\varphi_k)$ only by the sign; accordingly, the direction
of the angular momentum will be reversed ,i.e., $p_{\sigma_{k+1}}=-p_{\sigma_k}$.

This  stable running requires that  
 as $k \to \infty$, 
\beq \label{equ:goal1}
 \sigma_k-\varphi_k \to C_1 \mbox{ and } p_{\sigma_k} \to C_2 
\eeq 
for right stances  and
\beq \label{equ:goal2}
 \sigma_{k+1}-\varphi_{k+1} \to -C_1 \mbox{ and } p_{\sigma_{k+1}} \to -C_2 
\eeq 
for left stances where $C_1$ and $C_2$ are pre-selected constants. 

For right stances, we let
\beqn \label{equ:bodyconv1}
  \hspace{-0.5cm}&&\sigma_{k+1}-\varphi_{k+1}+C_1=(1-K_4)(
      \sigma_k-\varphi_k-C_1) \cr
 \hspace{-0.5cm}&&p_{\sigma_{k+1}}+C_2=(1-K_5)(p_{\sigma_k}-C_2)
\eeqn
where $0<K_4,K_5<1$.
For left stances, we may let 
\beqn \label{equ:bodyconv2}
 \hspace{-0.5cm}&& \sigma_{k+2}-\varphi_{k+2}-C_1=(1-K_4)(
      \sigma_{k+1}-\varphi_{k+1}+C_1) \cr
\hspace{-0.5cm} && p_{\sigma_{k+2}}-C_2=(1-K_5)(p_{\sigma_{k+1}}+C_2).
\eeqn

Note that $\varphi_k$ and $\varphi_{k+1}$ are known from (\ref{eq:phi1}) and (\ref{eq:phi2}) in Appendix \ref{apxa}.
It is not difficult to see that the discrete systems
(\ref{equ:bodyconv1}) and (\ref{equ:bodyconv2}) achieve the desired
convergence specified by (\ref{equ:goal1}) and (\ref{equ:goal2}).

Therefore, for the  $k$th stance, we want to design the control
torque $\tau$ so that starting from $p_{\sigma_k}$ and $\sigma_k$, the
runner will reach the state $p_{\sigma_{k+1}}$ and $\sigma_{k+1}$ given
by
\beqn \label{equ:ns}
    p_{\sigma_{k+1}}&=&(1-K_5) p_{\sigma_k}+(2-K_5)C_2 \cr
   \sigma_{k+1}&=&(1-K_4)\sigma_k+K_4 \varphi_k+ \cr
             && \pi -\phi_k-2\alpha_k+(2-K_4)C_1.
\eeqn

We  formulate an optimal control problem with the starting state given
by $(p_{\sigma_k},\sigma_k)$ and ending state given by
$(p_{\sigma_{k+1}},\sigma_{k+1})$ in equation (\ref{equ:ns}). The ending state should be achieved 
within the duration  $T_k$ for the $k$th stance  with the cost function $\int_0^{T_k} \tau^2(t)dt$ minimized.  
The solution for $\tau_k(t)$ for the $k$th stance  can be obtained
by applying the maximum principle. We have
\beq
  \tau_k(t) = \frac{1}{2}(A_{2_k}-\frac{A_{1_k}}{I}t) 
\eeq
where $t\in [0,T_k]$, $I$ is the moment of inertia, and 
\beqn \label{equ:ns2}
  A_{1_k}&=&\frac{24
    \,I^2}{T^3_k}(\sigma_{k+1}-\sigma_k)-
    \frac{12\,I}{T^2_k}(p_{\sigma_{k+1}}+p_{\sigma_k})\cr
\noalign{\medskip}
  A_{2_k}&=& \frac{12\,I}{T_k^2}(\sigma_{k+1}-\sigma_k)- 
     \frac{4}{T_k}p_{\sigma_{k+1}}-\frac{8}{T_k} p_{\sigma_k}\;.  
\eeqn
The detail of this derivation is included in Appendix \ref{apxb}. 
The duration $T_k$ can be computed in the similar way as $\phi_k$ in 
(\ref{equ:constraints}) as
\beq
 T_k = 2 \int_{\eta_k}^{\eta_{\rm min}}
       \frac{1}{\sqrt{2E-\frac{p_{\psi_k}^2}{\eta^2}-2V(\eta)}} d \eta.
\eeq

In equations (\ref{equ:ns}) and (\ref{equ:ns2}),  although $\tau_k(t)$ is open-loop control during each stance for $t\in [0,T_k]$, feedback is achieved through $p_{\sigma_k}$
and $\sigma_k$ when switching from one stance  to another.

\section{\label{sec:SIM}Simulations}

We present simulation results to demonstrate tracking a circle centered at the origin with
radius  $0.02$m. The parameters for the LLS model are the same as
in section \ref{sec:constraint}. The desired distance to the circle is $0.03$m. Initially, the runner 
start from $(0.1,0)$ outside the circle. The speed of the COM is $0.2$m/s. The initial direction of
the velocity is $\pi/3$. The leg placement angle $\alpha_i$ are constrained to be within the interval $(\pi/6,\pi/3)$.
We change the spring constant $b_k$ so that $q_i$ is always equal to $1.53$cm, which is $90\%$ of the leg 
length at rest.The gain $K_i$ is selected to be $0.5$. When $\theta_i$ can not be achieved by $\alpha_i$, 
we simply use the value for $\alpha_i$ that will minimize the differences between desired $\tilde \theta_i$ and
achievable $\theta_i$; hence implemented the approximation method.

\begin{figure}[htbp]
\begin{center}
\subfigure[trajectory of COM]{
\psfig{figure=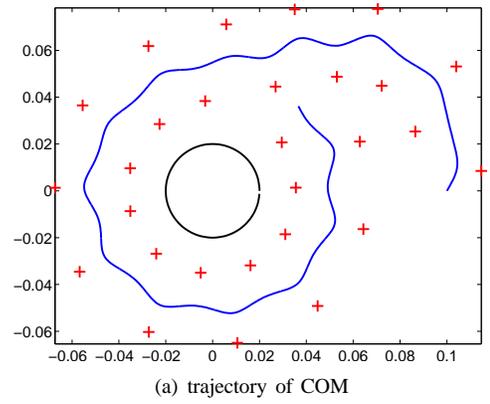,width=2.5in}
}
\subfigure[distance between COM and the circle]{
\psfig{figure=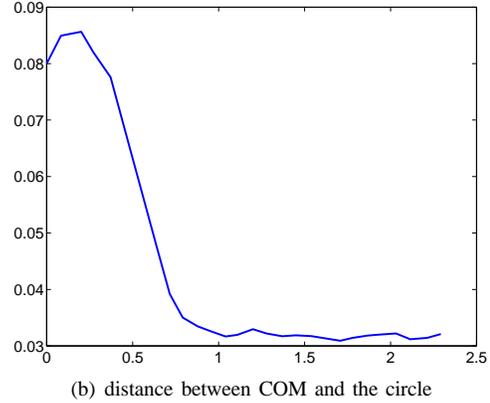,width=2.5in}
}
\end{center}
\caption{\it  Tracking a circle with radius $0.02$m. In (a), the cross symbols indicate the touchdown points. The units for both
horizontal and vertical axis are meters. In (b), the
distance (in meters) between the COM and the circle is plotted as a function of time (in seconds). We can see it converges to the desired separation $0.03$m. }
\label{fig:BugFigure10}
\end{figure}

The trajectory of the
COM and the distance between the COM and the circle are plotted in Figure \ref{fig:BugFigure10}.
We see that the convergence is achieved after $12$ stances which take less than one second. 

\section{Summary and Future Work}

We have analyzed the control of LLS model and   designed a  hybrid curve tracking
control law for legged locomotion. Using measurements of the curve for feedback,
 the discrete algorithm guarantees convergence to
the desired curve path. During each stance, the controlled continuous dynamics is analyzed.
The parameters of the LLS model is determined to implement the discrete algorithm at the
beginning of each stance. We have also investigated the effects of parameter constraints.
These constraints limited tracking ability. For straight lines and convex curves, a steady state
can be reached. The robustness of these steady states depends on the range for the parameters.

Interesting results regarding  wall following
behaviors of cockroaches are reported by Camhi and Johnson in
\cite{Camhi99}. 
When its antenna  touches the wall,
a cockroach turns away from the wall but keeps the antenna in contact
with the wall for a certain time period. The experiments there are performed near a raffled piecewise linear
wall, not a 
smooth curved wall.  Using a smooth curved wall with convex shape will also be interesting
 since a steady state can be reached.

Recently, a wall following wheeled robot using antenna like
tactile  sensor was reported in \cite{lamperski05}; curve tracking for atomic force microscope 
is discussed in \cite{Andersson05}; a general boundary tracking control law is derived for
Newtonian particles in \cite{ZhangPSKJusth04}.
Our work, although intended for legged locomotion, may
be adapted to handle other cases regarding curve tracking for platforms 
with hybrid motion dynamics.

\appendices
\section{\label{apxa}Discrete dynamics for leg parameters}
The feasible $\theta_k$ and $\theta_{k+1}$ that can be
generated from the LLS model parameters must satisfies
 (\ref{equ:alpha}) and (\ref{equ:alpha2}).  Equation (\ref{equ:alpha}) subtract (\ref{equ:alpha2}) yields
\beqn \label{equ:thetak1}
        \theta_{k+1} -\theta_k &=& (\zeta_{k+1}-\varphi_{k+1})-(\zeta_k-\varphi_k)+\pi-\cr
           && (\alpha_k+\alpha_{k+1})-\frac{\phi_k+\phi_{k+1}}{2}.
\eeqn
 The angle $(\zeta_k-\varphi_k)$ is measured between the COM velocity vector and the
tangent vector of the desired curve. In a similar fashion, we find
\beqn \label{equ:thetak2}
        \theta_{k+2} -\theta_{k+1} &=& (\zeta_{k+2}-\varphi_{k+2})-(\zeta_{k+1}-\varphi_{k+1})-\pi+\cr
           && (\alpha_{k+2}+\alpha_{k+1})+\frac{\phi_{k+2}+\phi_{k+1}}{2}.
\eeqn
We now establish a difference equation which describes the change of $(\zeta_k-\varphi_k)$. 

 Comparing  the COM velocity $\bv_k$ and $\bv_{k+1}$, we notice that they are reflectively symmetric to each other with $\bq_k$ 
as the axis of symmetry, as shown in Figure \ref{fig:BugFigure7}. This can be proved from the solution of equations (\ref{equ:LLS}).
From Figure \ref{fig:BugFigure3}, we conclude that
\beq \label{eq:phi1}
    \varphi_{k+1} - \varphi_k = 2 (\frac{\pi}{2}-\frac{\phi_k}{2}-\alpha_k)
\eeq
from a right stance to a left stance and
\beq \label{eq:phi2}
   \varphi_{k+2} - \varphi_{k+1} = -2 (\frac{\pi}{2}-\frac{\phi_{k+1}}{2}-\alpha_{k+1})
\eeq
from a left stance to a right stance.
From Figure \ref{fig:BugFigure2}, we can derive the change of $\zeta_k$ as
\beq \label{eq:zeta1}
       \zeta_{k+1} - \zeta_k = \gamma_k.
\eeq
The angle $\gamma_k$ can be determined using  the sine law
\beq
     \frac{\sin \gamma_k}{q_k} =\frac {\sin \beta}{\rho_{k+1}+\frac{1}{\kappa}}.
\eeq
We use $\beta =\pi/2 -\theta_k$ and $\rho_{k+1} = \rho_k + f_k$ to obtain
\beq
          \sin \gamma_k = \frac{q_k}{\lambda_k + f_k} \cos \theta_k.
\eeq
From (\ref{eq:phi1}) and (\ref{eq:zeta1}), we obtain
\beq \label{eq:zeta2}
           (\zeta_{k+1} -\varphi_{k+1})-(\zeta_k-\varphi_k) = \gamma_k - (\pi - \phi_k -2\alpha_k).
\eeq
Similarly,
\beq \label{eq:zeta3}
           (\zeta_{k+2} -\varphi_{k+2})-(\zeta_{k+1}-\varphi_{k+1}) = \gamma_{k+1} + (\pi - \phi_{k+1} -2\alpha_{k+1}).
\eeq

Using  (\ref{eq:zeta2}) and (\ref{equ:thetak1}),  we deduce that
\beq \label{equ:theta2}
  \theta_{k+1} -\theta_k = -(\alpha_{k+1} -\alpha_k) - \frac{\phi_{k+1}-\phi_k}{2} + \gamma_k.
\eeq 
Using (\ref{eq:zeta3}) and (\ref{equ:thetak2}), we have
\beq \label{equ:theta4}
  \theta_{k+2} -\theta_{k+1} = (\alpha_{k+2} -\alpha_{k+1}) +\frac{\phi_{k+2}-\phi_{k+1}}{2} + \gamma_{k+1}.
\eeq  
Equations (\ref{equ:theta2}) and (\ref{equ:theta4}) must be satisfied for all feasible $\theta_k$, $\theta_{k+1}$, and $\theta_{k+2}$.
 
\section{\label{apxb}Optimal control for rigid body dynamics}
The system equations for the rigid body dynamics is
\beqn
     \dot \sigma &=& \frac{p_\sigma}{I} \cr
     \dot p_\sigma &=& \tau.
\eeqn
We want to minimize the cost function $\int_0^{T_k} \tau^2(t) dt$. Applying
the maximum principle, we define the controlled Hamiltonian as
\beqn
    H_{\lambda,\tau}& = &\lambda_1 \dot \sigma + \lambda_2 \dot p_\sigma - \tau^2 \cr
&=& \lambda_1 \frac{p_\sigma}{I} + \lambda_2 
\tau -\tau^2
\eeqn
where $\lambda_1$ and $\lambda_2$ are the adjoint variables.
The optimal control $\tau$ that  minimizes $H_{\lambda,\tau}$ is computed by
letting $\partial{H_{\lambda,\tau}}/\partial{\tau}=0$. This yields
$\tau = \lambda_2/2$. The Hamiltonian for the system under this control
is 
\beq
    H_\lambda = \lambda_1 \frac{p_\sigma}{I} + \frac{1}{4}\lambda_2^2.
\eeq
This Hamiltonian induces the following Hamilton's equations:
\beqn
      \dot \sigma &=& \frac{p_\sigma}{I} \cr
     \dot p_\sigma &=& \frac{\lambda_2}{2} \cr
     \dot \lambda_1 &=& 0 \cr
     \dot \lambda_2 &=& - \frac{1}{I} \lambda_1.
\eeqn
From these equations we first observe that $\lambda_1(t)=\lambda_1(0)$.
Then we can solve for $\lambda_2$ as
\beq
          \lambda_2(t) = \lambda_2(0) -\frac{\lambda_1(0)}{I}t.
\eeq
In order to determine $\lambda_1(0)$ and $\lambda_2(0)$, we integrate
\beq
    \dot p_\sigma = \frac{1}{2} (\lambda_2(0)-\frac{\lambda_1(0)}{I} t)
\eeq
from $0$ to $T_k$. This yields
\beq \label{equ:psigma}
      p_{\sigma_{k+1}} = p_{\sigma_k}+ \frac{1}{2} \lambda_2(0) T_k - 
\frac{1}{4} \frac{\lambda_1(0)}{I} T_k^2. 
\eeq
We then integrate 
\beq
       \dot \sigma = \frac{p_\sigma}{I} =\frac{1}{I}\left( p_{\sigma_k}+ \frac{1}{2} \lambda_2(0) t -  \frac{1}{4} \frac{\lambda_1(0)}{I} t^2\right). 
\eeq
This gives us
\beq \label{equ:sigma}
       \sigma_{k+1} = \sigma_k + \frac{1}{I} p_{\sigma_k} T_k + \frac{1}{4I} \lambda_2(0) T_k^2 - \frac{1}{12 I^2} \lambda_1(0) T_k^3.
\eeq
From (\ref{equ:psigma}) and (\ref{equ:sigma}) we can solve for $\lambda_1(0)$ and $\lambda_2(0)$. One can verify that
 $\lambda_1(0)=A_{1_k}$ and $\lambda_2(0)=A_{2_k}$, where $A_{1_k}$ and $A_{2_k}$ are given in (\ref{equ:ns2}), are the
solutions.

\end{document}